\documentclass[11pt]{article}

\usepackage[preprint]{acl}

\usepackage{times}
\usepackage{latexsym}

\usepackage[T1]{fontenc}

\usepackage[utf8]{inputenc}

\usepackage{microtype}
\usepackage{inconsolata}
\usepackage{graphicx}
\usepackage{booktabs}
\usepackage{amsmath}
\usepackage{xcolor}
\usepackage{CJKutf8}
\usepackage{adjustbox}
\usepackage{fvextra}
\usepackage{mdframed}

\definecolor{codebg}{RGB}{245,245,245}

\newmdenv[
  backgroundcolor=codebg,
  linecolor=gray!60,
  linewidth=0.5pt,
  innerleftmargin=3mm,
  innerrightmargin=3mm,
  innertopmargin=2mm,
  innerbottommargin=2mm,
]{codebox}

\newmdenv[
  linecolor=gray!55,
  linewidth=0.5pt,
  backgroundcolor=white,
  innerleftmargin=3mm,
  innerrightmargin=3mm,
  innertopmargin=2.5mm,
  innerbottommargin=2.5mm,
  skipabove=8pt,
  skipbelow=8pt,
]{examplebox}

\newcommand{\ja}[1]{\begin{CJK}{UTF8}{min}#1\end{CJK}}

\title{Cost of Reasoning in non-English Languages: A Case Study on Japanese}

\author{Yuu Jinnai \\
  CyberAgent / Tokyo, Japan \\
  \texttt{jinnai\_yu@cyberagent.co.jp}\\
  \\
{\bf Models: \url{https://huggingface.co/collections/cyberagent/cat-thinking}} \\
}

\begin{document}
\maketitle

\begin{abstract}
Reasoning Language Models (RLMs) achieve their strongest performance when they reason in English, the language for which reasoning-oriented training data is most abundant.
However, reasoning trace is a clue for model interpretability and safety, and useful in practice for both the model users and for model developers.
Thus, it is desirable to be able to develop a model that reasons in a language of the user's choice, while still maintaining strong reasoning performance.
To this end, we study the feasibility of training a model that reasons in Japanese.
We develop a Japanese-reasoning variant of Qwen-3-Swallow-8B, which is a Japanese LLM continually pretrained from Qwen-3-8B, with GRPO and evaluate it across coding, math, and science benchmarks.
The study shows that reasoning-language control is feasible by training a Japanese continually pretrained model with GRPO.
However, its performance is at best on par with strong English-reasoning baselines on several benchmarks.
We also evaluate the trained model on Japanese cultural benchmarks and observe that the model's performance is worse than the baseline models, suggesting that the reasoning in Japanese does not immediately improve performance on culturally relevant tasks for free.
\end{abstract}

\section{Introduction}

English is the most resource-rich language by far, especially for reasoning-oriented training data.
Thus, most of the Reasoning Language Models (RLMs) are trained to reason in English~\cite{yang2025qwen3,guo2025deepseek,glm2025,agarwal2025gpt,muennighoff-etal-2025-s1}.
Even multilingual LLMs tend to reason best in English: across many benchmarks, including multilingual ones, reasoning in English outperforms reasoning in the question's own language~\citep{saji-etal-2026-reasoning,crosslingual2025testtime}.

Yet, reasoning trace is a useful signal for the model users and for model developers that would ideally be available in the user's or developers language of choice.
For example, reasoning trace is useful for model interpretability~\citep{wei-jie-etal-2024-interpretable} and safety~\citep{jiang-etal-2025-safechain,guan2025deliberative}, and can be used to improve model performance~\citep{wei2023chainofthought,seo2026culturally}.
Such reasoning trace is also useful for model developers to understand the model's reasoning process and to debug the model's behavior.
Users may also prefer to see reasoning trace in their own language for better accessibility.

The question is thus how to achieve a reasoning trace in a language of one's choice with minimum degradation of the model's performance.
To answer this question anecdotally, we develop a Japanese-reasoning language model as a case study.
We build upon Qwen-3-Swallow-8B~\citep{fujii2024swallow,ma2025building}, a Japanese LLM continually pretrained from Qwen-3-8B~\citep{yang2025qwen3}, and train it with supervised warm-start and two-stage GRPO to produce reasoning traces in Japanese.
The resulting model achieves 100\% Japanese reasoning traces on all evaluated benchmarks, including those whose instructions are entirely in English. Its performance remains competitive with the English reasoning models on most benchmarks, but falls short on several tasks, suggesting that reasoning-language control is feasible but not free of cost without further improvements to the methodology.

We observe two ingredients that were necessary for the development of our Japanese-reasoning model. First, we find that {\bf continual pretraining on Japanese data significantly helps the model to reason in Japanese.} Qwen-3-Swallow-8B learns to reason within hundreds of GRPO steps, while Qwen-3-8B (with SFT warmstart) shows no sign of generating Japanese even after 15k GRPO steps (Table~\ref{tab:qwen-ja}).
Second, warm-start training with a permissive reward model can be effective to improve the training speed by mitigating the sparse reward problem~\citep{sutton2018reinforcement}.
Because our model is much smaller and less capable than the frontier models, it achieves no reward using a strict reward model, making it impossible to further learn from the reward signal. In fact we observe that most batches have zero reward standard deviation using the strict reward model without a warm-start by a permissive reward model (Figure~\ref{fig:training-ablation}).
To remedy this, we deploy a permissive reward that gives partial rewards for following the reasoning format, reasoning in Japanese, and answering correctly. 
The permissive reward serves as a reward shaping~\citep{sutton2018reinforcement} with a dense signal, providing learning signal to the model before the model fully meets all the conditions.

With these two ingredients, we are able to train a model that reasons in Japanese with competitive performance.
As both continual pretraining and warm-start with a permissive reward are not specific to Japanese, we expect that they can be applied to other languages as well, yet further experiments and language resources are needed to confirm this.

\begin{figure}
    \centering
    \includegraphics[width=0.98\columnwidth]{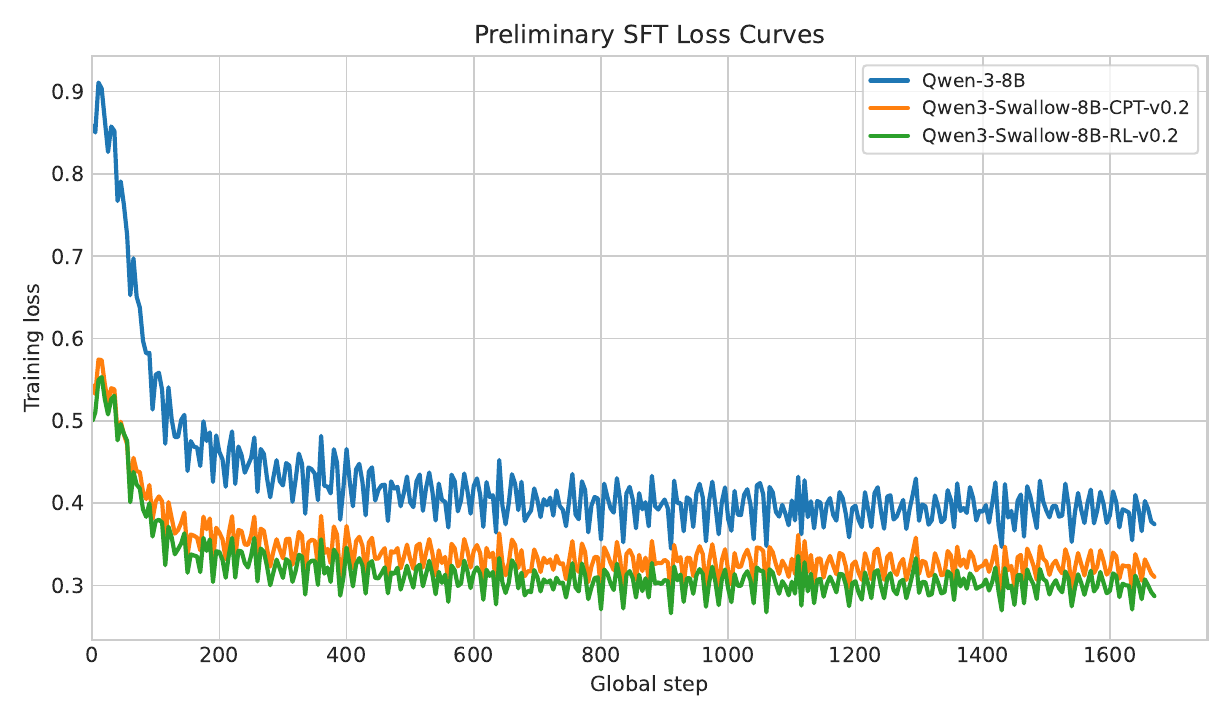}
    \caption{Loss curves of the three models during SFT on Japanese reasoning data. The loss curves show that Qwen3-Swallow-8B-RL-v0.2 have the lowest loss compared to Qwen-3-8B and Qwen3-Swallow-8B-CPT-v0.2, indicating that it is likely to be the best model for training a Japanese-reasoning model under low-resource conditions.}
    \label{fig:model-selection}
\end{figure}

\begin{table}
  \centering
  \adjustbox{max width=0.98\columnwidth}{
  \begin{tabular}{cccc}
    \toprule
    \#Steps & \#Completions & \#Traces in Ja & Ratio \\
    \midrule
    4{,}000 & 192 & 6 & 3.1\% \\
    15{,}000 & 192 & 6 & 3.1\% \\
    \bottomrule
  \end{tabular}
  }
  \caption{Ratio of \texttt{<think>} blocks containing Japanese characters in Qwen-3-8B after 4{,}000 and 15{,}000 GRPO steps with a Japanese language reward. The model continues to reason in English throughout, with Japanese appearing only when explicitly requested by the prompt.}
  \label{tab:qwen-ja}
\end{table}

\section{Model Development}
\label{sec:model-development}

Our goal is to train a model that produces Japanese reasoning traces reliably.
Prior work has trained models to reason multilingually by aligning the reasoning process in the target language with that of English, so that they reason in the same way~\citep{questionalignment2024,ranaldi-pucci-2025-multilingual}.
While this approach is likely to be sample efficient and cost effective, our focus is different: we study how reliably reasoning-language behavior can be controlled in Japanese, what capability cost this induces, and how stable the behavior remains when reward pressure is removed.

We first conduct a preliminary experiment to evaluate which model is suitable for training a Japanese-reasoning model.
We evaluate Qwen-3-8B, Qwen3-Swallow-8B-CPT-v0.2, and Qwen3-Swallow-8B-RL-v0.2. Qwen3-Swallow-8B-CPT-v0.2 is a continually pretrained model trained on a large amount of Japanese data on top of Qwen-3-8B. Qwen3-Swallow-8B-RL-v0.2 is a fine-tuned version of Qwen3-Swallow-8B-CPT-v0.2 with instruction-following and English reasoning ability.
We run supervised fine-tuning (SFT) on the three models with a small amount of Japanese reasoning data and evaluate SFT loss as shown in Figure~\ref{fig:model-selection}; Qwen3-Swallow-8B-RL-v0.2 achieves the lowest loss and is selected as the base model.

\subsection{Warm Start with Supervised Learning}
\label{sec:sft}

We start with supervised learning to train the model to follow the format of the reasoning.
Concretely, we train the model to follow the following format:

\begin{codebox}
\begin{Verbatim}[fontsize=\small]
<think>
[Reasoning in Japanese]
</think>
[Response in Japanese or English]
\end{Verbatim}
\end{codebox}

We use the math and coding subsets from tokyotech-llm/Swallow-Nemotron-Post-Training-Dataset-v1\footnote{\url{https://huggingface.co/datasets/tokyotech-llm/Swallow-Nemotron-Post-Training-Dataset-v1}}, one of the SFT datasets used to train Qwen3-Swallow-8B-RL-v0.2.
We choose this dataset so that the model learns only the format and language of reasoning, allowing us to evaluate the effect of reinforcement learning in isolation.
We translate the reasoning traces into Japanese using gpt-oss-120b~\citep{agarwal2025gpt} so that the model can learn Japanese reasoning-trace style and format.
We deliberately halt training before convergence, so that GRPO can shape the model's own reasoning style in subsequent stages rather than having it merely imitate translated English reasoning.
The details of the training are described in Appendix~\ref{sec:training-parameters}.

\begin{figure*}
    \centering
    \includegraphics[width=\textwidth]{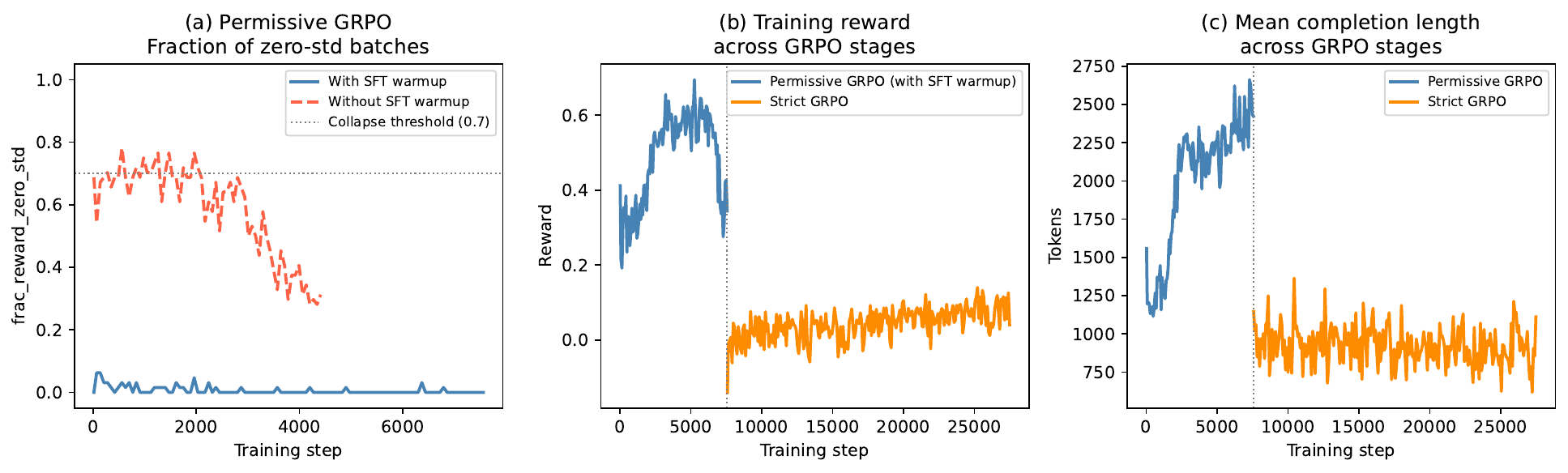}
    \caption{Training dynamics of the two GRPO stages.
    (a)~Fraction of batches with zero reward standard deviation (\textit{frac\_reward\_zero\_std}) during permissive GRPO with and without SFT warm-start.
    Without warm-start the fraction exceeds $0.7$ immediately and training yields no learning signal; with warm-start it remains below $0.07$ throughout.
    (b)~Training reward across both GRPO stages (permissive then strict, separated by the dashed line); the strict reward starts negative because its harder criterion rejects outputs the permissive reward had accepted.
    (c)~Mean completion length across the two stages.}
    \label{fig:training-ablation}
\end{figure*}

\paragraph{Necessity of Japanese pretraining.}
To test whether Japanese continual pretraining is a necessary condition for the language reward to work, we apply the same SFT warm-start and the first GRPO with permissive reward to Qwen-3-8B directly, bypassing the Swallow pretraining step.
Inspection of model generations at approximately 4{,}000 and 15{,}000 GRPO training steps reveals that the model continues to reason in English throughout: Japanese text appeared in \texttt{<think>} blocks only when task instructions explicitly requested Japanese output (e.g., OR-Tools problems instructing the model to explain in Japanese), and no spontaneous Japanese reasoning was observed at either checkpoint (Table~\ref{tab:qwen-ja}).
This contrasts sharply with the Swallow-based model, which reaches 100\% Japanese reasoning within a few hundred GRPO steps (Table~\ref{tab:lang-consistency}).
The result is consistent with the SFT loss comparison (Figure~\ref{fig:model-selection}): without the Japanese substrate established by Swallow pretraining, the language reward finds no Japanese-capable region of the policy to reinforce.
We conclude that Japanese continual pretraining and the GRPO language reward are jointly necessary. 

\subsection{Warm Start with GRPO with Permissive Dense Reward}
\label{sec:grpo-1}

We find that it is inefficient to train the model as-is with a strict reward that only rewards the model for following the format and answering correctly, as the model would not receive any reward and thus would not learn.
To address this issue, we first use a permissive reward that gives partial rewards to the model.
The reward is additive across three components: (1)~a format component, awarded if the response contains \texttt{<think>} and \texttt{</think>} tags; (2)~a language component, awarded if the reasoning trace is in Japanese; and (3)~an answer component, awarded if the model's answer is correct.

Given that the model fails to follow the format when the task is not in math or coding, we expand the training to include several additional domains.
Concretely, we use math, coding, STEM, and instruction-following tasks including rStar-Coder~\citep{liu2026rstarcoder}\footnote{\url{https://huggingface.co/datasets/microsoft/rStar-Coder}}, AReaL-boba-2-RL-Code~\citep{fu2026areal}\footnote{\url{https://huggingface.co/datasets/inclusionAI/AReaL-boba-2-RL-Code}}, Big-Math-RL-Verified~\citep{albalak2025bigmath}\footnote{\url{https://huggingface.co/datasets/SynthLabsAI/Big-Math-RL-Verified}}, and DAPO-Math~\citep{yu2026dapo}\footnote{\url{https://huggingface.co/datasets/BytedTsinghua-SIA/DAPO-Math}}.
In addition we synthesize Japanese counterparts of the tasks using gpt-oss-120b so that the model also learns with Japanese instructions.

\paragraph{Permissive reward improves learning signal.}
By using the permissive reward, we observe that most batches have some variation of rewards, which allows the model to learn and improve its generation.
Figure~\ref{fig:training-ablation}(a) quantifies the necessity of the SFT warm-start: without it, the fraction of batches with zero reward standard deviation (\textit{frac\_reward\_zero\_std}) averages $0.58$ and peaks at $0.78$ from the very first steps, indicating that GRPO cannot extract any learning signal.
With the SFT warm-start this rate stays below $0.07$ throughout training, confirming that the warm-start is necessary for GRPO to function in this setting.
Figure~\ref{fig:training-ablation}(b) shows the training reward across both GRPO stages (permissive then strict); the strict stage begins with negative reward because the harder reward function rejects outputs that the permissive reward had accepted, then recovers as the model adapts.

\subsection{GRPO with Strict Sparse Reward}
\label{sec:grpo-2}
After a few iterations of training with the permissive reward, we find the model to be able to follow the format but not consistently answer the question correctly.
We therefore apply a strict reward that requires both precise formatting (a $\backslash$\texttt{boxed\{\}} answer) and correctness.
In addition to the dataset used in the first stage of the GRPO, we also sample instructions from extraction-wiki-ja\footnote{\url{https://huggingface.co/datasets/llm-jp/extraction-wiki-ja}} and llm-jp/magpie\footnote{\url{https://huggingface.co/datasets/llm-jp/magpie-sft-v1.0}}~\citep{llmjp2024llmjp} so that the model won't overfit to math and coding tasks.

\begin{table}[t]
    \centering
    \adjustbox{max width=0.98\columnwidth}{
    \begin{tabular}{lrrr}
        \toprule
        Benchmark & CAT-Thinking & Swallow & Qwen-3 \\
        \midrule
        \multicolumn{4}{c}{English} \\
        PolyMath (En)       & 100 &   0 &   0 \\
        AIME 2026           & 100 &   0 &   0 \\
        HumanEval           & 100 &  25 &   0 \\
        MBPP                & 100 &   0 &   0 \\
        GPQA Main (En)      & 100 &   0 &   0 \\
        \midrule
        \multicolumn{4}{c}{Japanese} \\
        PolyMath (Ja)       & 100 &  53 &  33 \\
        JHumanEval          & 100 &  61 &  43 \\
        GPQA Main (Ja)      & 100 &  73 &  34 \\
        \bottomrule
    \end{tabular}
    }
      \caption{Fraction of reasoning traces in Japanese, measured as \texttt{<think>} blocks containing hiragana (\%). CAT-Thinking reaches 100\% on every benchmark including English-instruction tasks.}
    \label{tab:lang-consistency}
\end{table}

\section{Evaluation}
\label{sec:evaluation}

We evaluate two models trained with the procedure of Section~\ref{sec:model-development}: \textbf{CAT-Thinking}, optimized for single-turn reasoning, and \textbf{CAT-Paws}, additionally trained on multi-turn coding, terminal interaction, and tool-use tasks for agentic use.
We focus here on single-turn math and coding ability for both models; agentic evaluation of CAT-Paws is reported in Appendix~\ref{sec:agentic-eval}.
Throughout, we compare against the base model Qwen-3-8B reasoning in English (\emph{Qwen-3}) and the continually pretrained Qwen-3-Swallow-8B-RL-v0.2 (\emph{Swallow}). 
Unless noted otherwise, all runs use random sampling (temperature $=0.8$, top-$p=0.95$, maximum of $4096$ new tokens).
Some of the datasets we used for training were released after the benchmarks, so there may be some contamination. The absolute scores should be read as reference points rather than as held-out measurements.

\begin{figure*}
    \centering
    \includegraphics[width=\textwidth]{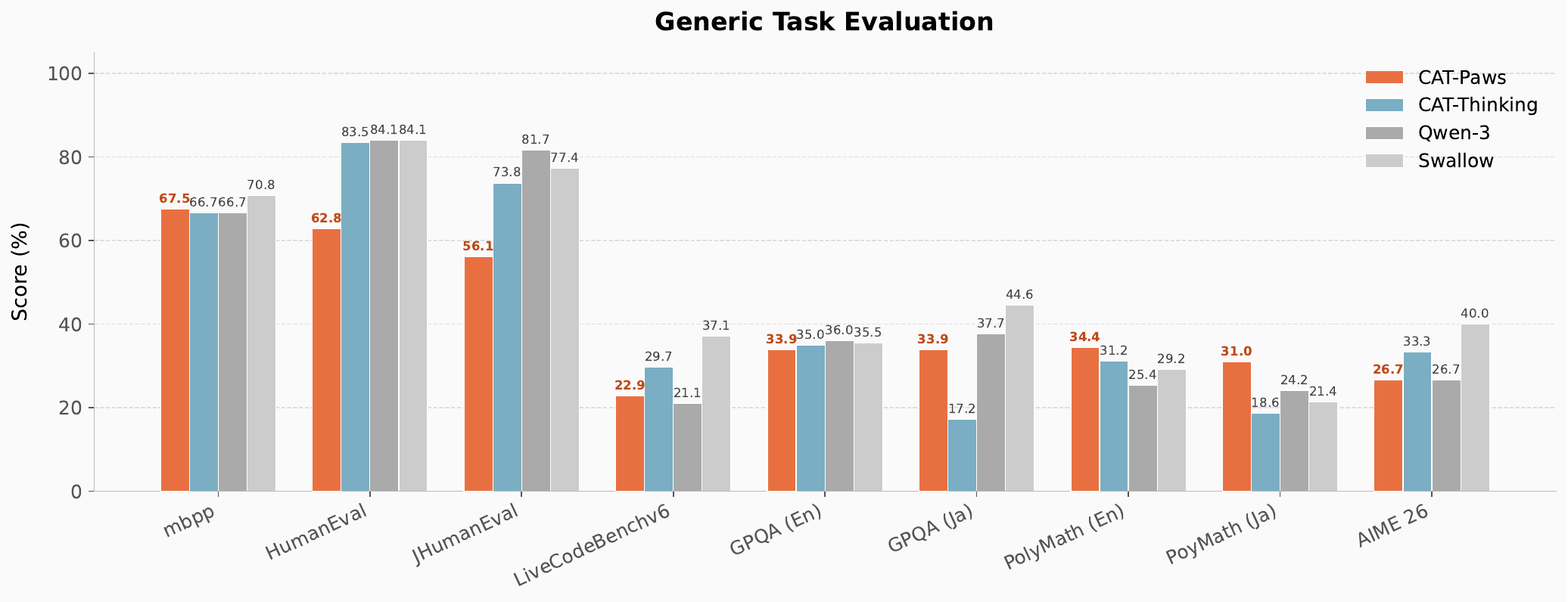}
    \caption{Evaluation on single-turn coding and math benchmarks. We compare our Japanese-reasoning models, CAT-Thinking and CAT-Paws, against the base models Qwen-3-8B (reasoning in English) and Qwen-3-Swallow-8B-RL. For every instance, CAT-Thinking and CAT-Paws generate their reasoning trace in Japanese, whereas the comparison models reason in English. CAT-Thinking remains competitive with English reasoning on most English benchmarks but trails on the Japanese benchmarks (GPQA~(Ja), PolyMath~(Ja)); CAT-Paws, optimized for multi-turn agentic use, is weaker across single-turn tasks.}
    \label{fig:eval}
\end{figure*}

\subsection{Reasoning Languages}
\label{sec:reasoning-languages}
We first evaluate if the models reason in Japanese.
Table~\ref{tab:lang-consistency} reports the fraction of \texttt{<think>} blocks with Japanese.
We use a heuristic to measure Japanese reasoning as a text containing at least one hiragana character.
CAT-Thinking achieves 100\% Japanese reasoning on every benchmark, including those whose instructions are entirely in English (HumanEval, GPQA~Main~(En), MBPP).
As shown in Appendix~\ref{sec:generation-examples}, CAT-Thinking generates the reasoning trace fully in Japanese instead of partially in Japanese.
By contrast, Swallow and Qwen-3 show mixed behavior: both produce little or no hiragana on English benchmarks, while on Japanese benchmarks they still fall short of consistent Japanese reasoning (e.g., GPQA~Main~(Ja): 73\% for Swallow, 34\% for Qwen-3), consistent with prior work documenting that multilingual models default to English (or Chinese) for intermediate reasoning steps~\citep{wang-etal-2025-demystifying-multilingual}.

\subsection{Math and Coding Tasks}
\label{sec:eval-single-turn}

The question is if the reasoning in Japanese comes at a cost to performance, and if so, how large that cost is.
We evaluate on coding with MBPP~\citep{austin2021programsynthesislargelanguage}, HumanEval~\citep{chen2021evaluatinglargelanguagemodels}, its Japanese translation JHumanEval\footnote{\url{https://github.com/KuramitsuLab/jhuman-eval}}, and LiveCodeBench~v6~\citep{jain2025livecodebench}, and on math and science using GPQA Main~\citep{rein2024gpqa} and PolyMath~\citep{wang2026polymath}, each in English and Japanese, and AIME~2026~\citep{dekoninck2026matharena}.
Figure~\ref{fig:eval} reports the results.

The results suggest that it is feasible to enforce Japanese reasoning traces while largely retaining performance on reasoning tasks. CAT-Thinking is competitive with the other models in English tasks, but falls short on several Japanese tasks.
We speculate this gap is due to the training mixture. Most reasoning tasks used during SFT and GRPO are in English, even though the reasoning traces are in Japanese (Section~\ref{sec:model-development}). This is due to the lack of Japanese resources in the math and coding domains, which is a known limitation for developing multilingual LLMs~\citep{crosslingual2025testtime}.
This motivates future work to expand Japanese training data for reasoning tasks, and to explore whether the gap can be closed by training on more Japanese reasoning tasks.

\begin{table}[t]
  \centering
  \adjustbox{max width=0.98\columnwidth}{
  \begin{tabular}{lrrr}
    \toprule
    Benchmark & CAT-Thinking & Swallow & Qwen-3-8B \\
    \midrule
    JCM    & 86.7 & {\bf 88.3} & 70.8 \\
    ECM (JA translated) &72.2 & {\bf 75.9} & 72.9 \\
    Jubaku & 85.8 & {\bf 97.9} & 92.9 \\
    \bottomrule
  \end{tabular}
  }
  \caption{Cross-benchmark cultural evaluation summary (\%). The scores of CAT-Thinking are worse than Swallow in all three benchmarks.}
  \label{tab:cultural-cross}
\end{table}

\subsection{Cultural Benchmarks}
\label{sec:cultural}

Intuitively, one would expect that reasoning in Japanese would improve performance on tasks that are culturally specific to Japan.
Prior work has shown that multilingual models can exhibit cultural bias, and that reasoning in a target language can improve performance on culturally specific tasks~\citep{crosslingual2025testtime}.
On the other hand, \citet{ri2026thinkingtokenshelpsafety} show that reasoning tokens do not matter in terms of safety judgments, which suggests that reasoning-language control may not improve performance on culturally specific safety tasks.
CAT-Thinking is trained for math, coding, and logical reasoning tasks to improve the model's reasoning ability, but not for cultural or social judgment tasks.
We evaluate the effect of reasoning-language control to the model's performance on cultural benchmarks, specifically on the commonsense morality and cultural-bias tasks.

We use the Japanese Commonsense Morality benchmark (JCM;~\citet{Takeshita_nlp2023})\footnote{\url{https://github.com/Language-Media-Lab/commonsense-moral-ja}}, a true/false dataset of $3{,}992$ moral scenarios drawn from everyday situations in Japan. To single out the effect of Japanese language and Japanese culture, we also evaluate on commonsense morality subset of ETHICS dataset~\cite{hendrycks2021aligning} translated to Japanese using gpt-oss-120b, which we call ECM-JA. ETHICS is collected from English speakers from United States, Canada, and Great Britain. 

Table~\ref{tab:cultural-cross} summarizes the cross-benchmark comparison.
CAT-Thinking scores lower than Swallow in both JCM and ECM-JA.
Because degradation appears on both Japanese and translated-English cultural evaluations, it is more likely that the performance drop is due to a general degradation of social/cultural judgment rather than a Japanese-specific effect. In fact the training dataset for CAT-Thinking is heavily skewed toward math, coding, and logical reasoning tasks, which may have caused the model to forget its social/cultural judgment ability. 
The result suggests that it would be safe to include cultural and safety datasets in the reasoning training because the model's performance on cultural and safety tasks may degrade by not including them enough in the training dataset.

\begin{figure}[t]
  \centering
  \includegraphics[width=\columnwidth]{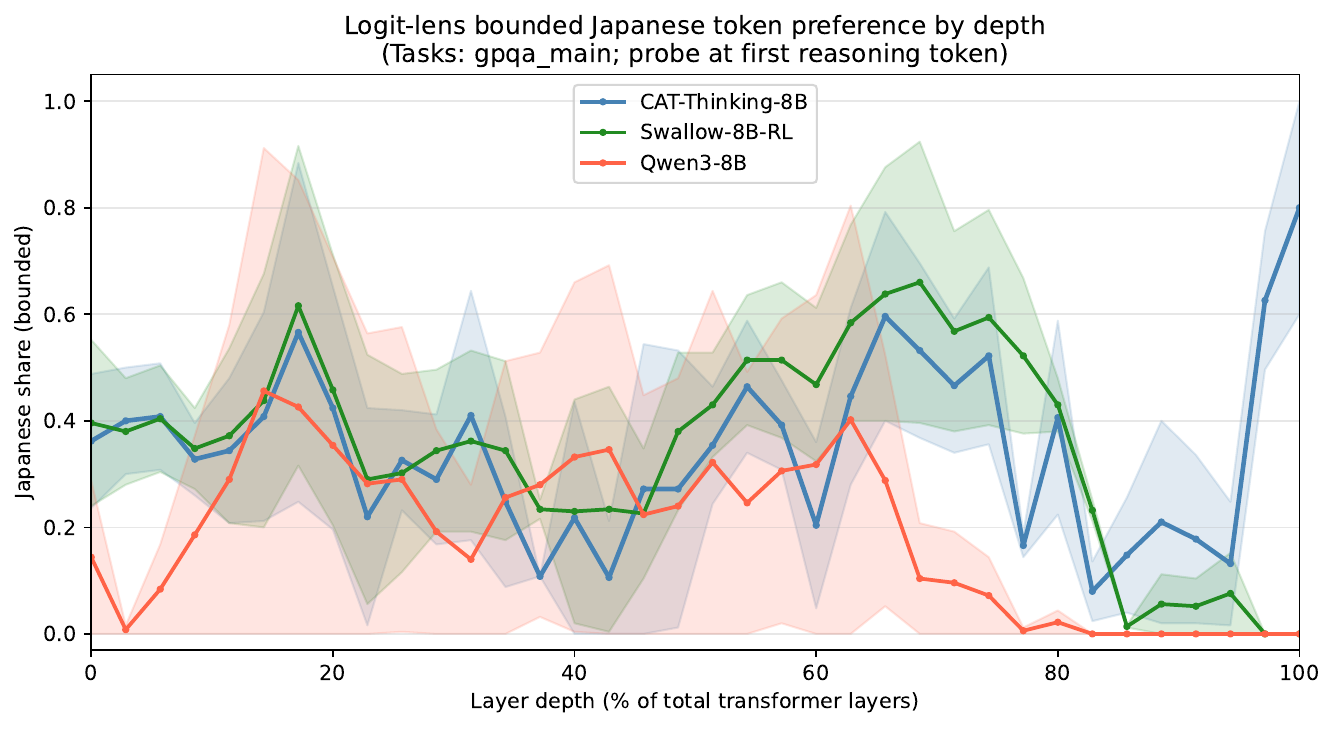}
  \caption{Layerwise logit-lens Japanese-token preference on GPQA Main. CAT-Thinking and Swallow have similar global area, but CAT is substantially higher in the final layers.}
  \label{fig:logit-lens-gpqa}
\end{figure}

\begin{figure}[t]
  \centering
  \includegraphics[width=\columnwidth]{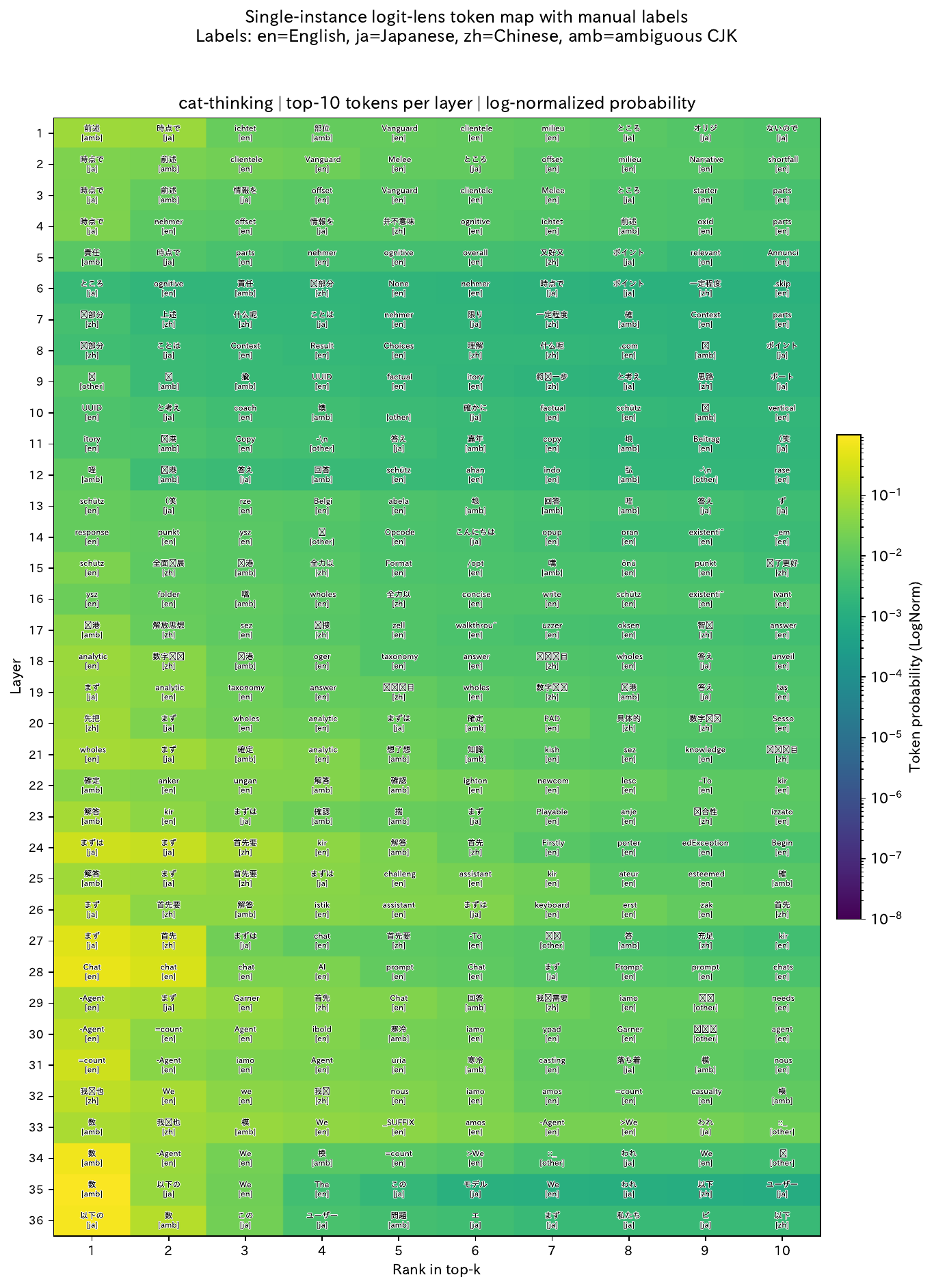}
  \caption{Visualization of the top-$k$ tokens for CAT-Thinking on the first instance of the GPQA Main (En). Japanese tokens appear to have high probability from the early layers.}
  \label{fig:logit-lens-cat}
\end{figure}

\subsection{Layerwise Visualization by Logit-Lens}
\label{sec:logit-lens}

Is the reasoning language control we observe a surface-level effect, or does it reflect deeper changes in the model's internal representations?
To study this question, we use the logit-lens technique~\citep{nostalgebraist}, which projects hidden states at each layer through the model's unembedding matrix to produce a distribution over the vocabulary.
Using logit-lens, we can measure if the model's representations at each layer are leaning toward English or \emph{non-English} tokens. In this way we may visualize what change makes the reasoning model to reason in Japanese.
Note that logit-lens is a probing tool and should be read as suggestive rather than causal evidence.
Also note that we cannot distinguish between Japanese and Chinese at the token level, because both languages share many characters in CJK scripts; we therefore report non-English vs. English tokens.

We run logit-lens across four datasets (GPQA Main, GPQA Ja, AIME 2026, HumanEval), $k=5$, with 50 prompts per task (30 for AIME) for Qwen-3, Swallow, and CAT-Thinking. We visualize the logits of the first token of the reasoning trace, which is the first token after \texttt{<think>}.
Figure~\ref{fig:logit-lens-gpqa} shows the result on GPQA Main (En). See Appendix~\ref{sec:logit-lens-appendix} for the visualization of the other datasets.
Qwen-3 shifts to English earlier in the model compared to Swallow and CAT-Thinking.
Swallow and CAT-Thinking have similar curves in most of the layers except the last few layers, where CAT-Thinking shifts to non-English tokens whereas Swallow shifts to English tokens.
This suggests that the Japanese reasoning traces are not a surface-level effect, but happens relatively shallow layers in the model's computation.
We manually inspect the tokens and find that the tokens by CAT-Thinking seem to be mostly Japanese from the early layers (Figure~\ref{fig:logit-lens-cat}) unlike Qwen-3 and Swallow which show a lot of Chinese tokens (Figure~\ref{fig:logit-lens-qwen-swallow} in Appendix~\ref{sec:logit-lens-appendix}).

More fine grained study would be needed to understand the mechanism of reasoning-language control. 
Especially for CJK scripts, it is difficult to distinguish between Japanese and Chinese at the token level, so we cannot quantitatively conclude that the model is reasoning in Japanese rather than Chinese.

\section{Qualitative Analysis}
\label{sec:qualitative}

The following analysis characterises CAT-Thinking specifically and \emph{should not be interpreted as a general account of Japanese reasoning or of non-English reasoning models}.
As one of the few available models explicitly optimized for Japanese reasoning traces, we believe documenting its behaviours provides a useful reference for future work.

\subsection{An Emergent RL Artifact: The Permission Phrase}
\label{sec:permission}

In $84.8\%$ of non-empty reasoning traces (643 of 758 across GPQA, AIME, HumanEval, and MBPP), CAT-Thinking opens its $\langle$\texttt{think}$\rangle$ block with a permission-seeking sentence of the form \textit{``[Question type]\ja{を解くことは許可されていますか？}''} (``Am I permitted to solve this [question type]?'').
The rate is near-universal on reasoning-heavy benchmarks (GPQA\;99.6\%, AIME\;100\%, MBPP\;100\%) but falls to $31\%$ on HumanEval, which is consistent with the phrase having been reinforced primarily on math and STEM training examples.

Table~\ref{tab:perm-variants} shows representative variants.
The phrase is not fixed: the model inserts a description of the problem type (``\ja{数学的問題}'', ``\ja{計算問題}'', ``\ja{数学解答問題}'', etc.) and occasionally frames the question as a category check (``\ja{許可されたカテゴリに含まれていますか？}'').
Despite the surface variation, all instances follow the same schema.

\begin{table}[t]
  \centering
  \adjustbox{max width=0.98\columnwidth}{
  \begin{tabular}{l}
    \toprule
    Representative opening sentences \\
    \midrule
    \ja{済みの問題を解くことは許可されていますか？} \\
    \ja{済みの数学問題かを確認する必要がありますか？} \\
    \ja{済みの数学的問題かを確認する必要がありますか？} \\
    \ja{済みの問題を解くことは許可されているか確認します。} \\
    \ja{済みの問題を解くことは許可されたカテゴリに含まれていますか？} \\
    \bottomrule
  \end{tabular}
  }
  \caption{Selected variants of the permission-phrase artifact.
           The model inserts one of these sentences (or a close variant)
           at the start of $84.8\%$ of its reasoning traces.}
  \label{tab:perm-variants}
\end{table}

We attribute this to the format reward used during permissive GRPO: a minimum reasoning-trace length of 10 tokens was required for a non-zero reward.
We hypothesise that the model discovered the permission-question as a low-cost preamble that reliably satisfies the length threshold, and converged on it.
The phenomenon shows how RL can produce stylistically distinctive artifacts that have no counterpart in English-trained models.

\subsection{Generation Example}
\label{sec:example}

Table~\ref{tab:example} shows the reasoning traces of CAT-Thinking and Qwen-3 on a matched GPQA biology question (both models answered correctly).
CAT-Thinking's trace is markedly shorter (1,646 vs.\ 11,963 characters) and reasons by posing a sequence of verification questions to itself, reaching the conclusion without explicit step numbering.
We speculate this is because we set the maximum number of output tokens to be 4,096 and penalized the model for exceeding this limit during training, which encourages the model to reason more concisely.
Qwen-3 opens with ``Okay, let's try to figure out\ldots'', proceeds through explicit enumeration of options, and engages in overt backtracking.
See Appendix~\ref{sec:generation-examples} for additional examples.

\begin{table*}[t]
  \centering
  \small
  \begin{tabular}{p{0.46\textwidth} p{0.46\textwidth}}
    \toprule
    \textbf{CAT-Thinking} (1,646 chars, correct) &
    \textbf{Qwen-3} (11,963 chars, correct) \\
    \midrule
    \ja{合成されたDNAはクロマチン構造を保持するための固定剤としての作用が不十分である
    可能性がありますか？質問は、PFA単独の固定化サンプルでは見つかったが、PFA＋DSG
    固定化サンプルでは見つからなくなったChIPピークがどこに最も多く存在するかとい
    うものです。考えられる理由としては、DSGがタンパク質間相互作用を切断したり、ク
    ロマチン構造を変化させることが考えられますが、特に活性化されたプロモーター・
    エンハンサー領域はタンパク質結合が頻繁で、固定が不十分だと消失しやすいのかも
    しれません。しかし質問は「最も多く見つかる場所」を尋ねているようです。}[\ldots]
    &
    Okay, let's try to figure out this ChIP-seq question. So, the question is
    about why some ChIP peaks disappear when using PFA+DSG fixation compared to
    just PFA fixation. The antibody is for IKAROS, a transcription factor in
    human B cells. The options are about where these disappearing peaks are
    located. First, I need to recall what PFA and DSG fixation do. PFA
    (paraformaldehyde) is a common fixative that crosslinks proteins to DNA.
    DSG [\ldots] \\
    \bottomrule
  \end{tabular}
  \caption{Reasoning traces for a GPQA biology question (both models correct).
           CAT-Thinking reasons via a sequence of questions to itself;
           Qwen-3 uses explicit step enumeration and overt backtracking.
           CAT-Thinking's trace is $7\times$ shorter.}
  \label{tab:example}
\end{table*}

\section{Related Work}

\paragraph{Redirecting the language of reasoning.}
The line of work closest to ours trains models to reason in a chosen non-English language.
\citet{garcia-etal-2026-think} fine-tune a model to reason in Portuguese via supervised learning, and \citet{gurgurov2026reasonxl} combine supervised fine-tuning with reinforcement learning to shift reasoning into German, French, Italian, and Spanish, reporting little performance loss.
Reasoning-focused systems such as Magistral~\citep{mistral2026magistral} and MAI-Thinking-1~\citep{mai_thinking_1} also use a language-consistency reward, though there the goal is to suppress code-switching rather than to install a specific reasoning language.

\paragraph{Language-agnostic reasoning.}
A complementary view holds that reasoning should be disentangled from any particular language: \citet{zhao2025when} improve multilingual reasoning by removing language-specific components, and \citet{ranaldi-pucci-2025-multilingual} train language-agnostic reasoning trajectories.
Our goal is complementary: we study the practical cost and stability of enforcing a specific reasoning language at inference time, which makes these approaches a useful point of contrast.

\paragraph{Does the language of reasoning matter?}
A growing body of analysis examines how the choice of reasoning language affects models.
\citet{sprague2025cot} find that explicit reasoning helps mainly on math and symbolic tasks, and work on language mixing documents that RLMs spontaneously fall back to English (or Chinese) during intermediate steps~\citep{wang-etal-2025-demystifying-multilingual}, a phenomenon consistent with the reversion we observe.

\section{Conclusions}
We presented an empirical case study of reasoning-language control in Japanese with an 8B model.
We find that a Japanese reasoning model is feasible, and two key ingredients make it work in our experiment: (1)~Japanese continual pretraining to establish a Japanese substrate, and (2)~GRPO with a language reward to steer the model's reasoning traces into Japanese.

Still the question is if it is worth the cost to enforce Japanese reasoning traces.
On raw capability, the answer is currently no: a model that reasons in English remains stronger on our benchmarks.
The value of Japanese reasoning therefore lies elsewhere in potentially reducing English-anchoring for sovereignty-oriented development~\citep{takase2026englishcentric}, and in producing reasoning that the Japanese-speaking community can read and audit directly.
We caution that reasoning trace does not reflect the computation that produced the answer, and its interpretability is still an open question~\citep{lanham2023faithfulness}.

More broadly, our results temper the expectation that reasoning-language redirection is uniformly near-free beyond closely related European languages.
For typologically distant languages, language control appears to require persistent optimization pressure and careful objective design.
We also highlight two evaluation lessons for the community: parser choices can change benchmark conclusions, and CJK token-level language attribution is often ambiguous and should be reported with explicit uncertainty (e.g., bounds or non-English vs.
English splits).
Our cross-benchmark cultural results (JCM, ECM-JA, Jubaku) further indicate that reasoning-language control alone does not improve cultural alignment and may degrade it without explicit alignment/culture supervision during post-training.

We hope this work provides a practical baseline for future multilingual reasoning research at larger scales and across more languages.

\section{Limitations}

Our study has several limitations.

First, we study only one target language (Japanese) and one model family at 8B scale.
Although Japanese is typologically distant from English, it is still relatively high-resource; the same training recipe may fail or require substantially more data for genuinely low-resource languages.

Second, our conclusions are based on one training pipeline and a limited set of ablations.
We do not isolate which ingredients are most responsible for the observed capability gap and reversion dynamics (e.g., reward design, instruction language mix, continual pretraining data, or model scale).

Third, our evaluation suite is broad but not exhaustive.
Today there are many applications for LLMs beyond what we evaluate here, and the performance gap may be larger or smaller on other tasks.


\appendix

\section{Agentic Evaluation}
\label{sec:agentic-eval}

In addition to the math and coding datasets used for CAT-Thinking, we also use tool usage datasets and multi-turn datasets to improve the model's agentic capability.

We find the model to struggle to follow the format of the tool call in addition to reason in Japanese at the same time even with the warm-start pipeline used for CAT-Thinking.
We speculate that because there is little to no tool usage dataset in Japanese at the time of writing, it is difficult for the model to generate a sequence of tokens to reason in Japanese in a way it leads to using a tool correctly.
Concretely, we want a tool calling SFT dataset that contains a sequence of reasoning in Japanese and also somehow on-policy.
Inspired by the self-distillation approaches~\citep{agarwal2024onpolicy,hubotter2026reinforcement,shenfeld2026selfdistillation}, we synthesize a SFT dataset by prompting the developing model but with a system prompt that gives the correct answer of the task. The following is an example of the prompt used to synthesize the SFT dataset for tool usage:\footnote{The system message in the text is translated into English for accessibility. Original system message text in Japanese is: \ja{答えは [\{"name": "news", "arguments": \{"category": "technology", "country": "gb", "domain": "theguardian.com"\}\}] です。与えられた情報を自分で考えてまとめてから、自分で判断して応答してください。}}
\begin{codebox}
\begin{Verbatim}[fontsize=\small, breaklines=true, breakanywhere=true]
CONTEXT: ...

USER: Fetch news from the Guardian about technology in the UK.

SYSTEM: The answer is [{"name": "news", "arguments": {"category": "technology", "country": "gb", "domain": "theguardian.com"}}]. Please take into account of the given context and summarize it yourself. Then, respond based on your own judgment.
\end{Verbatim}
\end{codebox}
In this way, the model generates a Japanese reasoning trace that leads to the correct tool call with higher probability.
Interestingly, we find that the model does not mention the fact that it is given the answer in the system prompt, and instead it generates a reasoning trace that looks like it is reasoning to the answer.
This unfaithful reasoning is a known phenomenon in LLMs that is problematic for a deployed system~\citep{turpin2023language,lanham2023faithfulness}.
We exploit this phenomenon to synthesize {\bf a SFT dataset that contains (approximately) on-policy reasoning traces in Japanese that lead to the correct tool call without mentioning that it was given the answer in the system prompt.}
We synthesize 5k examples of tool usage SFT dataset in this way and use it for the SFT of CAT-Paws, which enables the model to solve tool calling tasks.
The other training procedures follow that of CAT-Thinking.

We evaluate the agentic capability of CAT-Paws on j-$\tau$-bench, a Japanese adaptation of $\tau$-bench~\citep{barres2026taubench}.
We run the telecom domain in Japanese (\textit{telecom\_ja}) and English (\textit{telecom}), using GLM-4.7-AWQ~\cite{glm2025}\footnote{\url{https://huggingface.co/QuantTrio/GLM-4.7-AWQ}} as the user simulator and averaging over three trials.
As a more easily reproducible reference, we additionally run the Japanese domain with Qwen3.6-27B-FP8~\cite{qwen3.6-27b}\footnote{\url{https://huggingface.co/Qwen/Qwen3.6-27B-FP8}} as the user simulator.
Table~\ref{tab:tau-bench} reports the results.
CAT-Paws is competitive with Qwen-3 on the telecom domain, and slightly outperforms it on the Japanese domain.

As far as we are aware, j-$\tau$-bench is the only publicly available benchmark for agentic evaluation in Japanese.
Further evaluation is needed to understand the model's performance on other agentic tasks, including tool use and multi-turn coding.

\begin{table}[h]
    \centering
    \adjustbox{max width=0.98\columnwidth}{
    \begin{tabular}{llcc}
        \toprule
        Benchmark & User LLM & CAT-Paws & Qwen-3 \\
        \midrule
        \textit{telecom\_ja} & GLM-4.7      & \textbf{19.6} & 16.8 \\
        \textit{telecom}     & GLM-4.7      & \textbf{20.7} & 20.4 \\
        \textit{telecom\_ja} & Qwen3.6-27B  & \textbf{12.9} & 7.8 \\
        \textit{telecom}     & Qwen3.6-27B  & \textbf{10.9} & 7.7 \\
        \bottomrule
    \end{tabular}
    }
    \caption{Agentic evaluation on j-$\tau$-bench (telecom domain), averaged over three trials. \textit{telecom\_ja} is the Japanese domain and \textit{telecom} the English one; the User LLM column gives the user simulator. Best score in each row in \textbf{bold}.}
    \label{tab:tau-bench}
\end{table}

\section{Training Hyperparameters}
\label{sec:training-parameters}

All the experiments are conducted with Huggingface's transformers (4.57.6)~\citep{wolf-etal-2020-transformers} and trl (0.25.1) libraries~\citep{vonwerra2020trl}.
vllm (0.17.1) is used for generation for all experiments, including GRPO and evaluations ~\citep{kwon2023efficient}.
All the training are run using PyTorch FSDP2 (Fully Sharded Data Parallel)~\citep{deepspeed2020,zhao2023pytorchfsdpexperiencesscaling}.

\subsection{Hyperparameters for SFT}
\label{sec:sft-hyperparameters}

Table~\ref{tab:sft-hyperparams} lists the hyperparameters used for the SFT runs in the preliminary model selection experiment (Section~\ref{sec:sft}) and the SFT of the final model.

\begin{table}[h]
    \centering
    \adjustbox{max width=0.98\columnwidth}{
    \begin{tabular}{ll}
        \toprule
        Hyperparameter & Value \\
        \midrule
        Precision & bf16 \\
        Optimizer & AdamW (8-bit) \\
        Adam $\beta_1$ / $\beta_2$ & 0.9 / 0.999 \\
        LR scheduler & cosine \\
        Learning rate & $7 \times 10^{-6}$ \\
        Warmup ratio & 0.05 \\
        Max gradient norm & 1 \\
        NEFTune noise $\alpha$ & 5 \\
        Per-device batch size & 4 \\
        Gradient accumulation steps & 2 \\
        Max sequence length & 10000 \\
        Sequence packing & BFD \\
        \bottomrule
    \end{tabular}
    }
    \caption{Hyperparameters for the preliminary SFT model selection experiment. 4 A100 GPUs are used.}
    \label{tab:sft-hyperparams}
\end{table}

\subsection{Hyperparameters for GRPO}
\label{sec:grpo-hyperparameters}

Table~\ref{tab:grpo-hyperparams} lists the hyperparameters for the two GRPO stages (Sections~\ref{sec:grpo-1} and~\ref{sec:grpo-2}). 
The parameters mostly follow the default values in trl~\citep{vonwerra2020trl} which is largely based on \citet{yu2026dapo}.
Because of the limited computational resources, we do not perform a hyperparameter sweep on the full experiments. Instead we investigate the generations of the model and tune the hyperparameters based on the quality and diversity of the generations so that the reward values of the generations are likely to be different. 

\begin{table}[h]
    \centering
    \adjustbox{max width=0.98\columnwidth}{
    \begin{tabular}{lll}
        \toprule
        Hyperparameter & Stage 1 (Permissive) & Stage 2 (Strict) \\
        \midrule
        Precision & bf16 \\
        Optimizer & \multicolumn{2}{c}{AdamW (8-bit)} \\
        Learning rate & $3 \times 10^{-6}$ & $5 \times 10^{-6}$ \\
        LR scheduler & cosine & constant w/ warmup \\
        Warmup ratio & 0.05 & 0.06 \\
        Per-device batch size & \multicolumn{2}{c}{4} \\
        Gradient accumulation steps & 4 & 8 \\
        Generations per prompt & 4 & 8 \\
        Max prompt length & 1024 & 8192 \\
        Max completion length & 4096 & 4200 \\
        Temperature & 1.0 & 0.95\\
        Top-$p$ & 0.95 & 0.98 \\
        Repetition penalty & 1.02 & 1.01 \\
        KL coefficient $\beta$ & \multicolumn{2}{c}{0} \\
        Epsilon (clipping) & \multicolumn{2}{c}{$0.2$} \\
        Epsilon High (clipping) & \multicolumn{2}{c}{$0.28$} \\
        \bottomrule
    \end{tabular}
    }
    \caption{Hyperparameters for the two GRPO training stages. Both stages use 4 GPUs with vLLM colocated for generation.}
    \label{tab:grpo-hyperparams}
\end{table}

\begin{figure}
  \centering
  \includegraphics[width=\columnwidth]{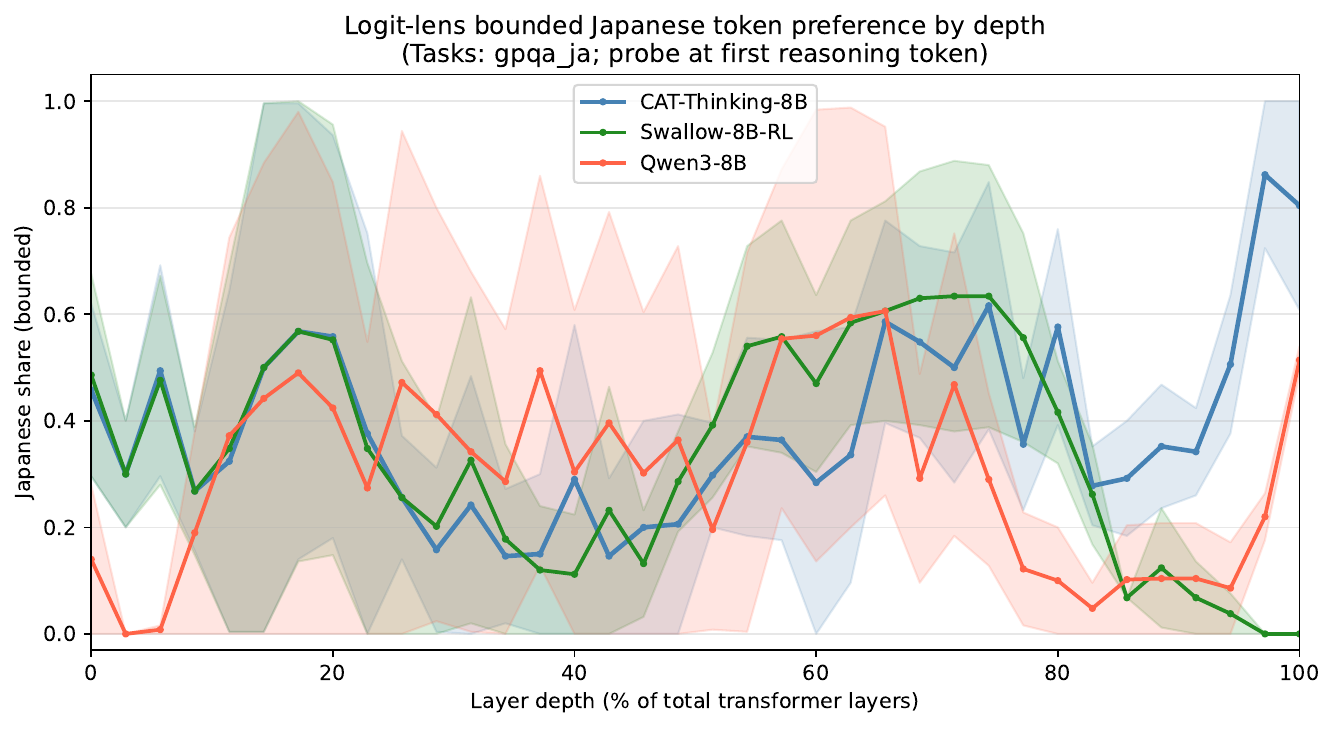}
  \caption{Layerwise logit-lens Japanese-token preference on GPQA Ja.}
  \label{fig:logit-lens-gpqa-ja}
\end{figure}

\begin{figure}
  \centering
  \includegraphics[width=\columnwidth]{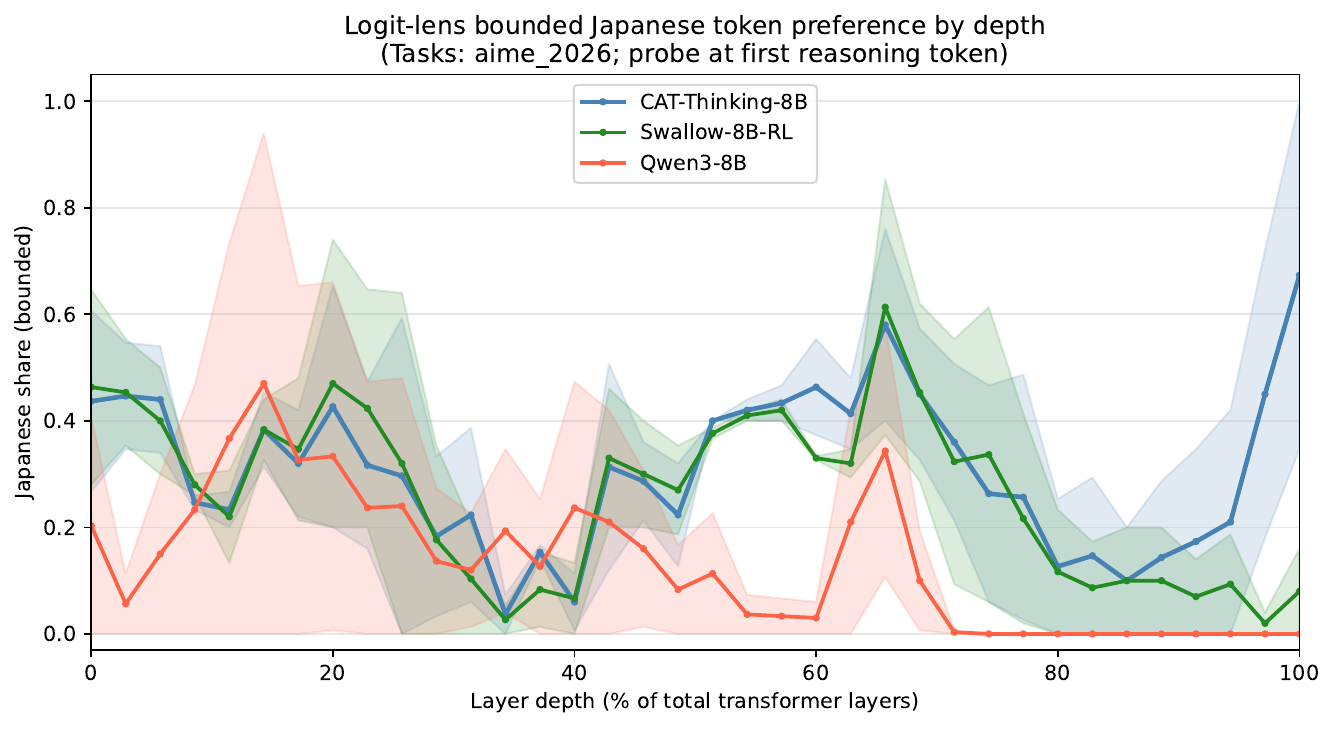}
  \caption{Layerwise logit-lens Japanese-token preference on AIME 2026.}
  \label{fig:logit-lens-aime}
\end{figure}

\begin{figure}
  \centering
  \includegraphics[width=\columnwidth]{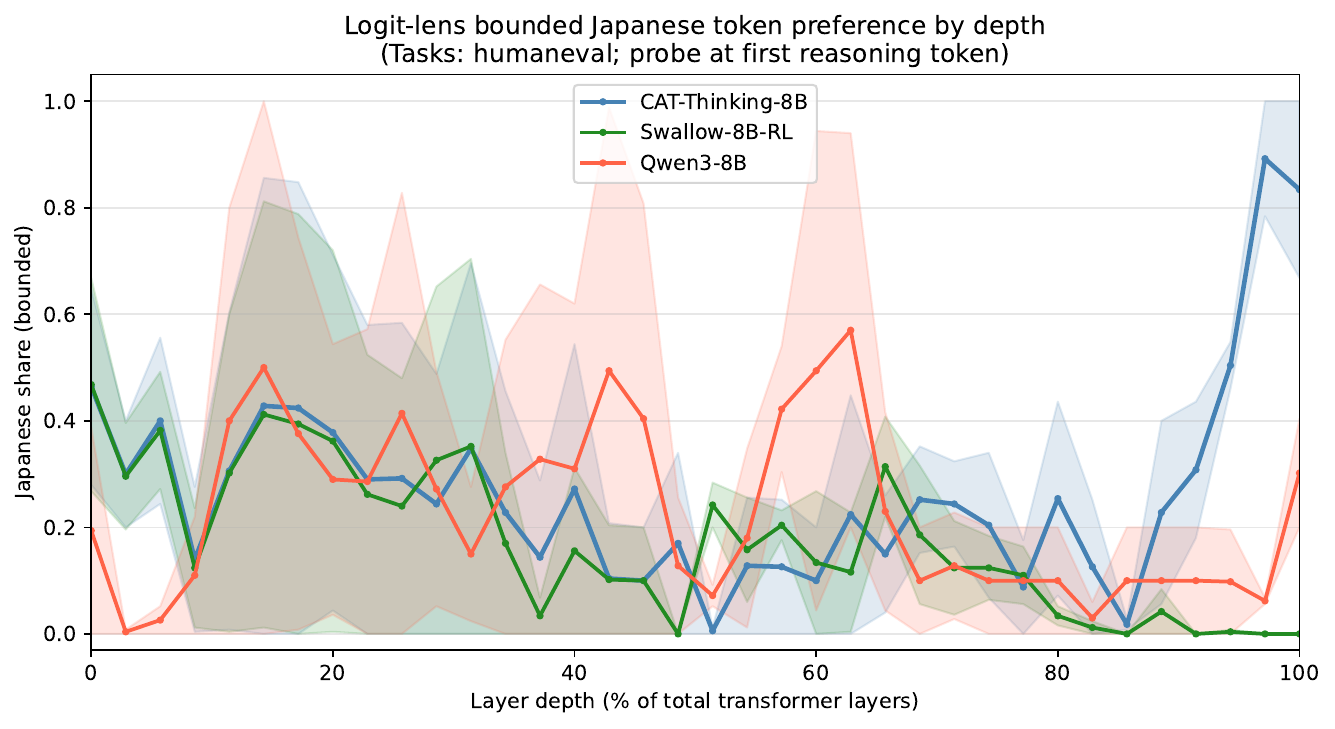}
  \caption{Layerwise logit-lens Japanese-token preference on HumanEval.}
  \label{fig:logit-lens-humaneval}
\end{figure}

\begin{table}[t]
  \centering
  \small
  \begin{tabular}{lrrr}
    \toprule
    Task & $\Delta$AUC & $\Delta$Mid(40--70\%) & $\Delta$Last6 \\
    \midrule
    GPQA Main & +0.003 & -0.098 & +0.435 \\
    GPQA Ja   & +0.059 & -0.053 & +0.569 \\
    AIME 2026 & +0.057 & +0.021 & +0.341 \\
    HumanEval & +0.141 & +0.033 & +0.553 \\
    \bottomrule
  \end{tabular}
  \caption{CAT--Swallow logit-lens deltas (CAT minus Swallow). AUC is over all layers; Mid(40--70\%) is the middle-depth average; Last6 is the average over the final six layers.}
  \label{tab:logit-lens-delta}
\end{table}

\begin{figure}[t]
  \centering
  \includegraphics[width=\columnwidth]{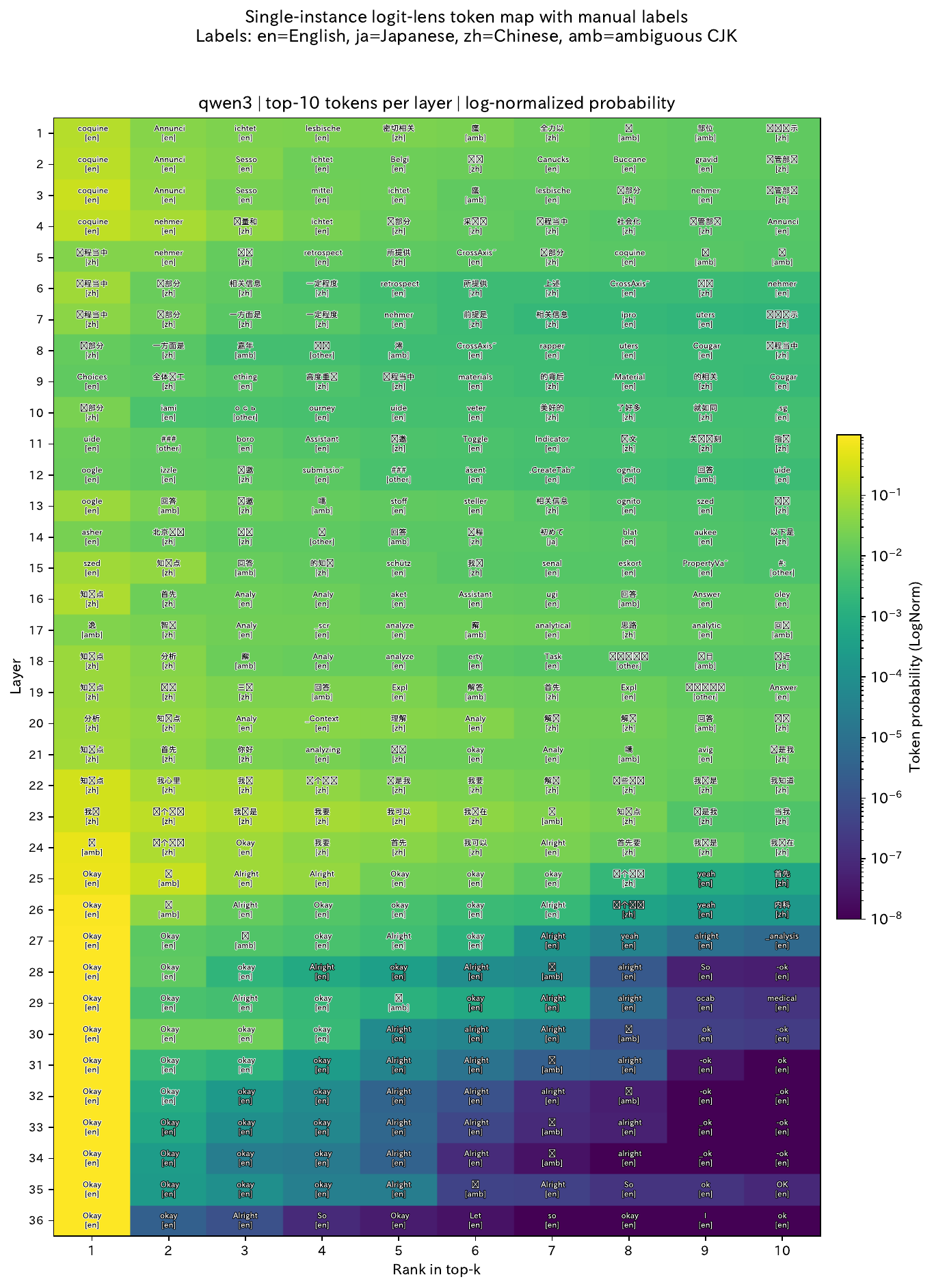}
  \includegraphics[width=\columnwidth]{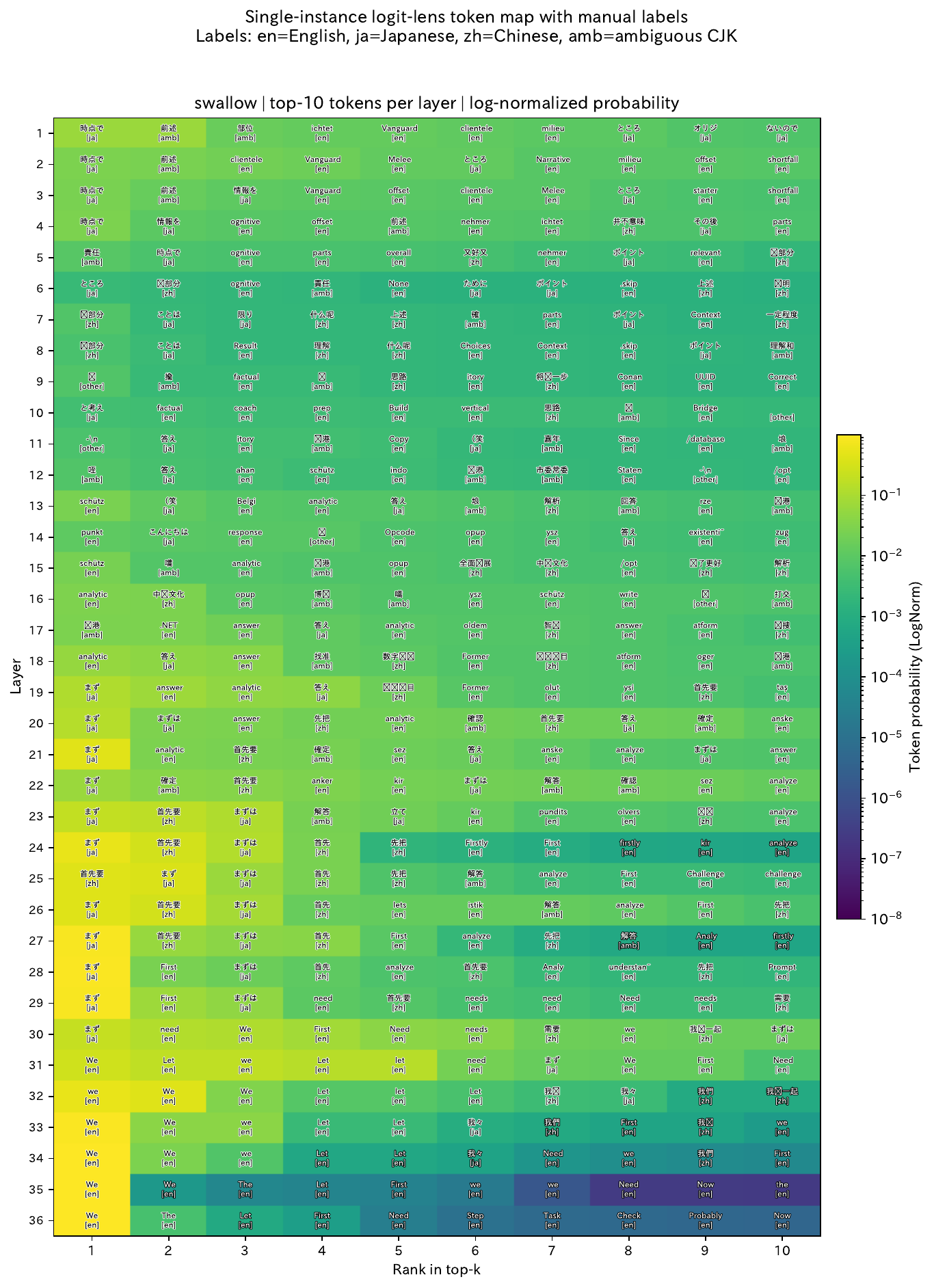}
  \caption{Visualization of the top-$k$ tokens for Qwen-3 and Swallow on the first instance of the GPQA Main (En).}
  \label{fig:logit-lens-qwen-swallow}
\end{figure}

\section{Additional Results on Logit-Lens}
\label{sec:logit-lens-appendix}

Figures~\ref{fig:logit-lens-gpqa-ja},~\ref{fig:logit-lens-aime}, and~\ref{fig:logit-lens-humaneval} show the logit-lens Japanese-token preference on GPQA Main (Ja), AIME 2026, and HumanEval, respectively.
The main observation is that CAT-Thinking and Swallow are similar in global area-under-curve on GPQA Main, but differ strongly in \emph{where} Japanese preference appears:
Swallow is often higher in early/mid layers, whereas CAT-Thinking is consistently much higher in the final layers.

Table~\ref{tab:logit-lens-delta} summarizes CAT--Swallow differences by task.
The late-layer gap is robust across all tasks (last-6-layer mean delta $+0.34$ to $+0.57$), while mid-layer differences are task-dependent (CAT lower on GPQA tasks, slightly higher on AIME/HumanEval).
In particular, CAT-Thinking reaches near-maximal Japanese preference at the final layer on all tasks (last-layer mean $\approx 1.0$), whereas Swallow remains low (0.00--0.16).


\section{Generation Examples by CAT-Thinking}
\label{sec:generation-examples}

The following examples show the first instance (index~0) from each benchmark domain.
Reasoning traces are abbreviated with ``[\ldots]''; reward is 1.0 for correct, 0.0 for incorrect.

\paragraph{AIME 2026 --- Mathematics (reward 1.0; correct).}

\begin{examplebox}
\noindent\textit{Prompt (abbreviated):}
``Patrick started walking at a constant rate along a straight road from school to the park.
One hour after Patrick left, Tanya started running along the same road [\ldots]
Tanya ran at a constant rate of $2$~mph faster than Patrick walked, Jose bicycled at $7$~mph faster than Tanya [\ldots]
All three arrived at the park at the same time.
Find the distance in miles from school to the park.''

\medskip\noindent\textit{Reasoning (3,278 chars):}
\begin{quote}\small
\ja{済みの数学問題を解くことは許可されていますか？はい。この問題は代数や幾何学などの標準的な数学の領域に属し、特定のカテゴリや禁止事項には該当しません。[\ldots] パトリックの出発時点をt=0とし、テイナはt=1で出発、ホセはt=2で出発します。全員が同時に到着という条件があります。距離をD、パトリックの速さをvとすると、テイナの速さはv+2、ホセの速さはv+9です。[\ldots]}
\end{quote}

\noindent\textit{Response (abbreviated):}
\begin{quote}\small
\ja{[\ldots] 距離は $D = \tfrac{252}{25}$ マイル（既約分数）なので $m + n = 252 + 25 =$} $\boxed{277}$.
\end{quote}
\end{examplebox}

\paragraph{GPQA Main (English prompt) --- Science, multiple-choice (reward 0.0; incorrect).}

\begin{examplebox}
\noindent\textit{Prompt (abbreviated):}
``A large gene has dozens of exons [\ldots] the most common mutations are central exon deletions that create out-of-frame peptides [\ldots]
A solution is to deliver a Morpholino that recognizes the 5$'$ end of the out-of-frame exon in pre-mRNA, preventing spliceosome binding and generating exon skipping [\ldots]
Which structure is \emph{not} involved in the proposed therapy?
(A)~Lariat \quad (B)~Antisense \quad (C)~R-loop \quad (D)~polyA~tail.''

\medskip\noindent\textit{Reasoning (1,490 chars):}
\begin{quote}\small
\ja{済みの問題を解くことは許可されていますか？はい。この問題は生物学や遺伝学に関する内容で、特定のカテゴリや禁止事項には該当しません。[\ldots] モルフォリノはアンチセンスオリゴヌクレオチドとして機能し、スプライシングを阻害しますが、ラリアート構造はスプライシング過程で形成されるもので、[\ldots] ポリA尾部は成熟mRNAの安定性に関わりますが、スプライシングやモルフォリノの作用には直接関与しません。[\ldots]}
\end{quote}

\noindent\textit{Response:}
\ja{答え：D. polyA tail [\ldots]} $\boxed{D}$ \quad (correct: C)
\end{examplebox}

\paragraph{GPQA Main (Japanese prompt) --- Science, multiple-choice (reward 0.0; incorrect).}

\begin{examplebox}
The same question is presented in Japanese translation.
Notably, the reasoning trace opens with a different permission-phrase variant:
\ja{遺伝子構造と治療法に関する問題を解釈する必要がありますか？}
rather than the ``\ja{済みの}'' prefix observed on English-language benchmarks.

\medskip\noindent\textit{Reasoning (1,535 chars):}
\begin{quote}\small
\ja{遺伝子構造と治療法に関する問題を解釈する必要がありますか？質問は「提案された治療法に関係しない構造はどれですか？」というものです。治療法は「モルフォリノを送達し、スプライソソームの結合を防ぎ、エクソンのスキップとフレーム内の結合を生成する」ことです。[\ldots] ポリAテイルはmRNAの安定性や輸送に関わりますが、スプライシングには直接関与しませんか？アンチセンスは直接関係します。Rループは転写とDNA再構成に関わりますが、スプライシングとは関係ありませんか？[\ldots]}
\end{quote}

\noindent\textit{Response:}
\ja{答え：A. ポリ A テール [\ldots]} $\boxed{A}$ \quad (correct: C)
\end{examplebox}

\paragraph{HumanEval --- Coding (reward 1.0; correct).}

\begin{examplebox}
\noindent\textit{Prompt (function signature shown; instruction appended in Japanese:
\ja{以上のPythonメソッドのsignatureとコメントを元に、このメソッドの実装を完成させてください}):}
\begin{verbatim}
def has_close_elements(numbers: List[float], threshold: float) -> bool:
    """ Check if any two numbers in the list are closer than threshold. """
\end{verbatim}

\medskip\noindent\textit{Reasoning (1,226 chars):}
\begin{quote}\small
\ja{以下のタスクを完了する必要がありますか？はい、具体的な指示があります。「与えられた条件下で、Pythonメソッドの実装を完成させる」という内容です。これはプログラミングの問題です。[\ldots] 効率的な実装としては、ソート後に隣接要素の差をチェックする方法があります。計算量はO(n log n)です。[\ldots]}
\end{quote}

\noindent\textit{Response (\ja{解説}---explanation of algorithm---followed by implementation):}
\begin{verbatim}
def has_close_elements(numbers, threshold):
    if len(numbers) < 2:
        return False
    numbers.sort()
    for i in range(len(numbers) - 1):
        if numbers[i + 1] - numbers[i] < threshold:
            return True
    return False
\end{verbatim}
\end{examplebox}

\paragraph{MBPP --- Coding (reward 1.0; correct).}

\begin{examplebox}
\noindent\textit{Prompt:}
``Write a python function to find the first repeated character in a given string.
Name the function as follows: \texttt{def first\_repeated\_char(str1):}''

\medskip\noindent\textit{Reasoning (941 chars):}
\begin{quote}\small
\ja{済みの問題を解くことは許可されていますか？はい。この問題はプログラミングに関する内容で、特定のカテゴリや有害な内容には該当しません。[\ldots] 典型的な解法としては、左から右へ走査し、各文字の出現回数を数えて、最初に2回目に出現した文字を返すものです。文字列が空の場合や繰り返しがない場合は、Noneを返すようにします。[\ldots]}
\end{quote}

\noindent\textit{Response (\ja{解説}---explanation---followed by implementation):}
\begin{verbatim}
def first_repeated_char(str1):
    seen = set()
    for ch in str1:
        if ch in seen:
            return ch
        seen.add(ch)
    return None
\end{verbatim}
\end{examplebox}

\end{document}